\documentclass[sigconf,screen]{acmart}

\renewcommand\footnotetextcopyrightpermission[1]{}
\usepackage{algorithmic}
\usepackage{graphicx}
\usepackage{textcomp}
\usepackage{xcolor}
\usepackage{multicol}
\usepackage{multirow}
\usepackage{booktabs}
\usepackage{bbm}
\usepackage{algorithm}
\usepackage{subcaption}
\usepackage{amsmath}
\usepackage{paralist}

\DeclareMathOperator*{\argmin}{arg\,min}
\DeclareMathOperator*{\argmax}{arg\,max}

\newcommand{\name}{DECO}

\AtBeginDocument{
  \providecommand\BibTeX{{
    \normalfont B\kern-0.5em{\scshape i\kern-0.25em b}\kern-0.8em\TeX}}}

\begin{document}

\title{Enabling On-Device Learning via Experience Replay with Efficient Dataset Condensation}

\author{Gelei Xu}
\affiliation{
 \institution{University of Notre Dame}
 \city{South Bend}
 \state{IN}
 \country{USA}
 }
 \email{gxu4@nd.edu}

\author{Ningzhi Tang}
\affiliation{
 \institution{University of Notre Dame}
 \city{South Bend}
 \state{IN}
 \country{USA}
 }
 \email{ntang@nd.edu}

 \author{Jun Xia}
\affiliation{
 \institution{University of Notre Dame}
 \city{South Bend}
 \state{IN}
 \country{USA}
 }
 \email{jxia4@nd.edu}

 \author{Wei Jin}
\affiliation{
 \institution{Emory University}
 \city{Atlanta}
 \state{GA}
 \country{USA}
 }
 \email{wei.jin@emory.edu}

 \author{Yiyu Shi}
\affiliation{
 \institution{University of Notre Dame}
 \city{South Bend}
 \state{IN}
 \country{USA}
 }
 \email{yshi4@nd.edu}

\begin{abstract}
  
Upon deployment to edge devices, it is often desirable for a model to further learn from streaming data to improve accuracy. However, extracting representative features from such data is challenging because it is typically unlabeled, non-independent and identically distributed (non-i.i.d), and is seen only once. To mitigate this issue, a common strategy is to maintain a small data buffer on the edge device to hold the most representative data for further learning. As most data is either never stored or quickly discarded, identifying the most representative data to avoid significant information loss becomes critical. In this paper, we propose an on-device framework that addresses this issue by condensing incoming data into more informative samples. Specifically, to effectively handle unlabeled incoming data, we propose a pseudo-labeling technique designed for unlabeled on-device learning environments. Additionally, we develop a dataset condensation technique that only requires little computation resources. To counteract the effects of noisy labels during the condensation process, we further utilize a contrastive learning objective to improve the purity of class data within the buffer. Our empirical results indicate substantial improvements over existing methods, particularly when buffer capacity is severely restricted. For instance, with a buffer capacity of just one sample per class, our method achieves an accuracy that outperforms the best existing baseline by 58.4\% on the CIFAR-10 dataset.
\end{abstract}

% \keywords{On-device learning, data-efficient learning, dataset condensation}

\maketitle

\section{Introduction}
Deep learning models have seen extensive application in edge devices, such as robots used in search operations \cite{shabbir2018survey} and UAVs for wildfire surveillance \cite{samaras2019deep}. Traditionally, these models are pre-trained on high-performance servers and deployed to edge devices without subsequent updates. However, it is often beneficial to continually update these models post-deployment, especially in unfamiliar environments where they need to adapt to new conditions. 

In practical deployments, edge devices often face non-stationary learning environments due to volatile data streams and limited storage capabilities. On-device learning typically processes unlabeled, non-i.i.d data that arrives in a stream and may exhibit temporary correlations. Moreover, the data distribution can shift over time. Consider autonomous driving as an example: a vehicle's camera might capture a series of images that are consecutive shots of another vehicle, followed by multiple images of the same traffic sign. Furthermore, due to the limited storage capacities of edge devices, data streams cannot be permanently stored and can be seen only once. Learning under these conditions can lead to an issue known as catastrophic forgetting \cite{mccloskey1989catastrophic, kemker2018measuring, aleixo2023catastrophic}, where the model loses previously acquired knowledge when new data arrives.

Replay-based strategies are commonly employed in on-device learning to mitigate catastrophic forgetting by maintaining a limited-size buffer that stores a selection of past samples~\cite{wu2021enabling, chaudhry2019tiny, aggarwal2023chameleon}. This buffer is used for rehearsal; as new data arrives, it is partially refreshed by replacing some older samples with new ones. The model is then trained on this continually updated buffer, which helps it retain information from earlier data, leading to improved training outcomes. Most replay-based strategies employ corset selection methods that retain the most representative samples in the buffer, while replacing others with new arriving data~\cite{vitter1985random,aljundi2019gradient}. However, these methods are often less effective. First, each data sample typically contains a low density of information, and the extremely limited storage capacity of the buffer further restricts the total amount of information it can hold. Second, the buffer must continually remove old samples to make room for new samples, and these discarded samples are often valuable. Consequently, over time, a significant amount of useful information may be lost, not only due to the suboptimal use of data before its replacement but also because of the forgetting of previously learned information. Given these challenges, an important question arises: How can we enhance the information density of the buffer while also preserving the information from old data when new data arrives?

Inspired by recent advancements in dataset condensation~\cite{wang2018dataset,zhao2020dataset,zhao2023dataset}, we propose that condensing incoming data into the buffer without removing any existing data could be a viable solution for on-device learning. Dataset condensation is designed to create a small, synthetic dataset learned from a large, original dataset, that provides sufficient information required for model training~\cite{wang2018dataset}. Utilizing such a condensed dataset enables the model to achieve performance comparable to that obtained when training on the full original dataset. For instance, a model can reach a 97.4\% test accuracy on the MNIST dataset using just 100 synthetic samples, which is close to the 99.6\% achieved with the full set of 60,000 images~\cite{zhao2020dataset}. Employing dataset condensation techniques to manage a buffer means the stored images are not limited to just the incoming data. Instead, it provides the flexibility to condense the representative features of incoming data into the existing buffer. Consequently, this approach can enhance the buffer's information density and more effectively mitigate the issue of catastrophic forgetting.

\textbf{Challenges in On-Device Learning.} While dataset condensation offers a potential solution for on-device learning, it also presents several challenges. \textbf{Firstly}, the effectiveness of current dataset condensation methods depends on the availability of label information. However, streaming data is often unlabeled, as it is impractical to label the data in real-time shortly after it is captured by sensors (e.g., cameras). Therefore, an effective and rapid labeling strategy is necessary. \textbf{Secondly}, even though the data can be appropriately labeled and condensed into the corresponding class of the buffer, label noise impacts the quality of the condensed dataset. Incorrect labeling of incoming data leads to erroneous condensation into an incorrect class, thereby reducing the deployed model's learning performance. \textbf{Finally}, most current dataset condensation methods were originally designed for offline learning settings. They often employ bi-level optimization, which requires multiple iterations of model updates before the synthetic data is updated. This iterative process, while effective in offline settings, is time-consuming and computationally demanding, making its application challenging in on-device learning scenarios.

In light of these challenges, we present a framework for on-device learning that updates synthetic data in a limited-size buffer using an efficient dataset condensation technique. When new data arrives, it is initially assigned pseudo-labels and filtered through majority voting. Subsequently, several techniques have been designed to optimize efficiency and effectively condense the data into the buffer. Additionally, to reduce the effects of inaccurate labels and improve the quality of the synthetic data, we use contrastive learning to enhance class purity within the buffer. Our results demonstrate significantly improved performance compared with existing methods. Our contributions can be summarized as follows:
\begin{compactenum}[(a)]
    \item To the best of our knowledge, we are the first ones to design the dataset condensation technique for buffer updates in on-device learning settings.
    \item We introduce a simple yet effective labeling technique to tackle the challenge of missing labels in on-device learning environments. Additionally, we design a contrastive learning approach to mitigate the negative impact of incorrect pseudo-labels.
    \item We optimize the algorithm to significantly speed up the condensation process, making it feasible for on-device settings without compromising accuracy.
    \item Extensive experiments have shown that our method significantly outperforms existing methods. For example, with a buffer capacity limited to just one sample per class, our method significantly outperforms the best existing baseline, achieving an accuracy improvement of 58.4\% on the CIFAR-10 dataset~\cite{krizhevsky2009learning}.
\end{compactenum}
\section{Background and Related Work}
\subsection{On-Device Learning}

On-device learning addresses the challenges of learning from an ongoing data stream on devices with limited memory resources~\cite{sangermano2022sample, wu2021enabling}. In this process, data may be presented in a non-i.i.d. manner, and each data sample is seen only once \cite{hayes2019memory}. Due to the continuously changing distribution of data streams, models often face the challenge of catastrophic forgetting. Replay-based strategies have proven effective in mitigating these issues by storing a selected subset of samples for later review. \cite{rebuffi2017icarl} proposes maintaining a set of representative examples per class, chosen to closely represent the class averages in the feature space. \cite{buzzega2020dark} focus on preserving previous knowledge by aligning new predictions with past logits. \cite{wu2021enabling} leverage contrastive learning features to evaluate the significance of individual samples. Additionally, \cite{de2021continual} integrates a nearest-mean classifier with an efficient reservoir sampling approach to enhance learning continuity.

Although these methods can be effective by selecting the most representative samples to store in the buffer, the information density of the buffer remains low due to the limited information each sample contains. Furthermore, the buffer must continuously remove old samples to accommodate new incoming data. Consequently, over time, a substantial amount of information is lost, as the majority of data is not fully utilized before being replaced. To address this challenge, it is crucial to develop a method that can increase the information density of the buffer while preserving the information from old data when new data arrives.

\subsection{Dataset Condensation}

Dataset condensation is designed to create a smaller, synthetic dataset from a larger training dataset. Training a model on this condensed dataset aims to achieve results comparable to those obtained from the original dataset. The concept of dataset condensation was originally proposed by \cite{wang2018dataset}, framing the condensation process as a learning-to-learn problem. Subsequent studies have focused on matching selected knowledge during network training, such as feature distributions~\cite{zhao2023dataset, wang2022cafe}, gradients, and training trajectories~\cite{cazenavette2022dataset, guo2023towards}. Building on this foundation, several approaches have been developed with subtle variations \cite{zhao2021dataset, lee2022dataset, kim2022dataset, wang2022cafe}. While several condensation techniques are available, in this paper, we focus on \textit{gradient matching} due to its intuitive approach and strong performance \cite{zhao2020dataset}. Note that similar to gradient matching, other dataset condensation techniques are also designed for offline use and face the same challenges such as lack of labels and high computational costs. Therefore, the method proposed in our paper can be flexibly adapted to other dataset condensation techniques as well.

The core idea of gradient matching is to replicate the training trajectory of the original dataset. Specifically, this technique aims to minimize the difference between the model's gradients with respect to the real and synthetic data at each training epoch. By doing so, the model $\theta$ optimized on the synthetic dataset will closely match those obtained from the original dataset. Let $\theta_t$ denote the model parameters at the $t$-th epoch, $\mathcal{S}$ as the synthetic dataset, and $\mathcal{R}$ as the original dataset. The gradient matching process can be formulated as follows:
\begin{gather}
\argmin_{\mathcal{S}} \sum_{t=0}^{T-1} \mathcal{D}(\nabla_{\theta} \mathcal{L}_{\theta_t}(\mathcal{R}), \nabla_{\theta} \mathcal{L}_{\theta_t}(\mathcal{S})), \nonumber \\
\text{s.t.} \quad \theta_{t+1} = \text{opt}_\theta(\theta_t, \mathcal{S}),
\label{eq:opt1}
\end{gather}
where $\mathcal{D}$ is the distance metric, $T$ represents the total number of training epochs, and $\text{opt}_{\theta}$ is the optimization algorithm applied to the loss function $\mathcal{L}_{\theta}$. 
\section{Dataset Condensation for On-Device Learning}
\label{sec:motivation}

\begin{figure*}[h]
    \centering
    \includegraphics[width=1\textwidth]{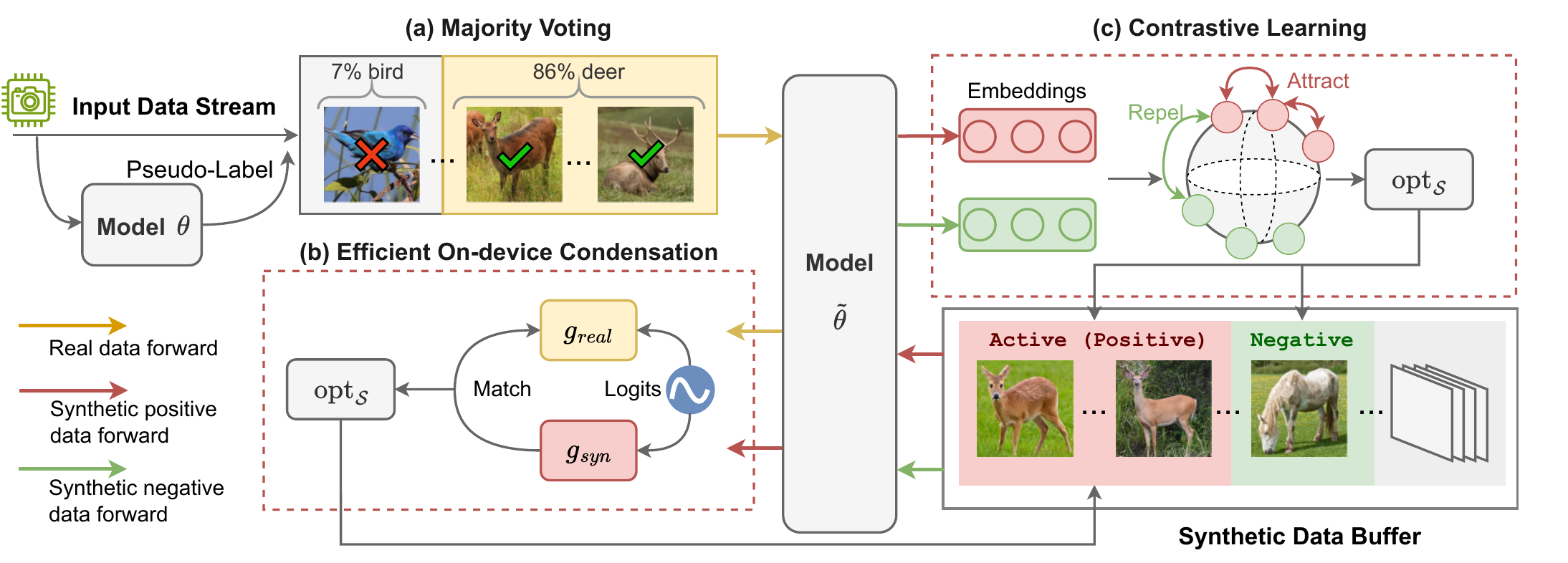}
    \caption{Overview of \name{}. The process begins by labeling and filtering the incoming unlabeled data stream through (a) \textit{majority voting}. Subsequently, the limited-size synthetic data buffer is updated through (b) \textit{efficient on-device condensation} and (c) \textit{contrastive learning} among the synthetic data within the buffer.}
    \label{fig:overview}
\end{figure*}

Before delving into our design goals and challenges, we first introduce some basic notations used in this paper.
After deploying the model $\theta$ on the edge device, it continuously learns from the input data stream $\mathcal{I}$ received on the fly. Note that the data instances within $\mathcal{I}$ are not labeled, yet they have potential category candidates $\mathcal{C}$. We maintain a limited-size data buffer on the edge device to store the condensed dataset $\mathcal{S} = \{(\mathbf{x}_i', y_i')\}$. The synthetic data instances $(\mathbf{x}_i', y_i')$ are evenly distributed across classes to ensure class balance. That is, for each class $c \in \mathcal{C}$, the number of instances $|\{(\mathbf{x}_i', y_i') | (\mathbf{x}_i', y_i') \in \mathcal{S}, y_i' = c\}|$ equals $|\mathcal{S}| / |\mathcal{C}|$.

\textbf{Our Design Goal.} Our goal is to enable the deployed model $\theta$ to continually learn from unlabeled input streaming data $\mathcal{I}$ while minimizing the forgetting of previously acquired knowledge~\cite{wang2024comprehensive}. Therefore, the overall workflow of on-device learning through dataset condensation can be summarized in two stages: (1) storing knowledge from the input data stream $\mathcal{I}$ into a condensed dataset $\mathcal{S}$ through gradient matching technique, and (2) the deployed model $\theta$ continually learns from $\mathcal{S}$ after several rounds of condensation.

\textbf{Obstacles in On-Device Learning.} Ideally, gradient matching in Eq.~\eqref{eq:opt1} aligns the training dynamics of the synthetic dataset with those of the real dataset, thereby preserving the knowledge learned from the original data and improving the model's performance. However, implementing this workflow into on-device learning presents several obstacles. Below we describe the obstacles encountered when applying this approach to on-device learning scenarios, as well as how they motivate our proposed solution:

\begin{compactenum}[(a)]
    \item \textit{\underline{Unlabeled Data.}} To the best of our knowledge, all current dataset condensation methods require labels for updates. Label information enables the synthetic data to be class-specific (for example, using real data of dogs to update corresponding synthetic data of dogs), making it more interpretable and easier for the model to learn. However, in on-device settings, data arrives in real-time and is typically unlabeled. Since it is impractical for humans to label these data quickly, it motivates us to generate pseudo-labels for these streaming data before condensation.
    \item \textit{\underline{Extensive Computational Cost.}} As shown in Eq.~\eqref{eq:opt1}, the conventional gradient matching scheme is a two-level optimization problem that involves updating synthetic images \(\mathcal{S}\) in the outer loop and optimizing network parameters \(\theta\) in the inner loop. Additionally, updating the synthetic data requires computing second-order gradients of $\mathcal{S}$ with respect to the distance metric $\mathcal{D}$, which necessitates square-level time and space computational complexity. These factors make the optimization process time-consuming and space-intensive, which is less practical for the limited computational resources available on edge devices.
    \item \textit{\underline{Inaccurate Buffer Updates.}} When updating synthetic data, assigning incorrect pseudo-labels to new incoming data can result in condensing the original data into the wrong class. Such mislabeling can increase the similarity of synthetic data across different classes, particularly those that are more alike (e.g., horse and deer), thereby reducing the quality of the condensed dataset in the buffer.
\end{compactenum}

\section{Methodology}

To address the obstacles discussed in Section~\ref{sec:motivation}, we propose a new framework, \name{}\footnote{\name{} is the acronym for \textbf{D}evice-centric \textbf{E}fficient \textbf{C}ondensation \textbf{O}ptimization.}, that utilizes the dataset condensation technique for on-device learning. This framework aims to generate pseudo-labels for new data while maximizing accuracy, reducing the impact of label noise on synthetic data, and improving the efficiency of gradient matching. In this paper, we consider image classification as a representative task for on-device learning to demonstrate its effectiveness~\cite{shabbir2018survey, samaras2019deep}. 

\subsection{Proposed Framework}

Fig.~\ref{fig:overview} provides a framework for our methodology. The model is pre-trained on a small amount of labeled data before deployment on the edge device, and the buffer is initialized with data that are condensed using such labeled data in offline settings. As the on-device learning begins and new segments of streaming data arrive, the system assigns pseudo-labels and filters the data using (a) \textit{majority voting} (Section~\ref{sec:labeling}). Next, the data of active classes is condensed to these corresponding synthetic samples through (b) \textit{efficient on-device condensation }(Section~\ref{sec:condensation}). Specifically, we update the buffer's synthetic samples by matching their gradients with the gradients of labeled incoming data. To reduce the impact of mislabeling, (c) \textit{contrastive learning} is applied to enhance the purity of the classes (Section~\ref{sec:contrastive_learning}). 
The model is updated every $\beta$ step to continuously learn from the input data stream. This iterative process aims to enhance the performance and adaptability of the on-device model to real-world data after deployment.

\subsection{Majority Voting based Pseudo-Label Assignment}\label{sec:labeling}

Suppose $\mathcal{I}_t = \{\mathbf{x}_i\}$ represents the $t$-th segment of input data stream from $\mathcal{I}$, where $\mathbf{x}_i$ is the $i$-th instance in $\mathcal{I}_t$ and is unlabeled. We use the deployed model $\theta$ to generate pseudo-labels for these instances, i.e., $\hat{y}_i = \argmax_c p_{\theta}(\mathbf{x}_i)c$, where $p_{\theta}(\mathbf{x}_i)_c$ is the predicted probability of class $c$ for input $\mathbf{x}_i$. Note that the deployed model is usually pre-trained in offline settings and aims to continue learning after deployment. This straightforward method leverages the deployed model's existing knowledge while facilitating immediate integration into the training process.

However, a fundamental issue with this method is the bidirectional influence between the quality of pseudo-labels and the performance of the deployed model. On one hand, the limited accuracy of pre-trained models leads to the generation of low-quality pseudo-labels; on the other hand, training with incorrect pseudo-labels may further decrease the model's accuracy. This leads to a detrimental cycle and potential training failure. Thus, it is necessary to develop a methodology that maintains relatively high predictive accuracy in pseudo-labels.

To ensure the accuracy of pseudo-labels, we recognize the non-i.i.d. nature of the input streaming data. As mentioned in the introduction, streaming data are often temporally correlated, exhibiting long sequences of data belonging to the same class. Therefore, we can infer that within a certain timeframe, the data received are more likely to belong to the same class. For example, if a majority of images are assigned the pseudo-label ``deer'' within a short timeframe, it is highly likely that a deer actually appeared during that period. Conversely, if there are very few images labeled as ``truck'' or ``boat'' among those labeled ``deer,'' there is a high probability that these few images are mislabeled.

Thus, based on the characteristics of the data stream, we propose a simple yet effective majority voting method to filter the samples with low confidence in pseudo-labels. After the model assigns pseudo-labels $\hat{y}_i$ to each data instance $\mathbf{x}_i$ in segment $\mathcal{I}_t$, we maintain a sliding window for it. Although any size of the sliding window can be used, for ease of description, we set it to the same size as the segment, i.e., $|\mathcal{I}_t|$. Subsequently, for each class $c \in \mathcal{C}$, we count the occurrences of pseudo-labels $\hat{y}_i$ across all samples within this sliding window. We then define the active classes $\mathcal{C}_t^A$ in this window as those whose count of pseudo-labels exceeds a certain threshold $M$:
\begin{equation}
    \mathcal{C}_t^A = \{c \in \mathcal{C} | \sum_{i=1}^{|\mathcal{I}_t|} \mathbbm{1}\left(\hat{y}_i = c \right) > M\}.
    \label{eq:filter}
\end{equation}

After the counting, we identify the active classes $\mathcal{C}_t^A$ within the current sliding window. This also implies that within this segment, the model will only update the synthetic data whose labels belong to $\mathcal{C}_t^A$ through gradient matching. We finally obtain the active data instances in $\mathcal{I}_t$ with their pseudo-labels, as well as the active subset of synthetic data, denoted as:
\begin{gather}
    \mathcal{I}_t^A = \{(\mathbf{x}_i, \hat{y}_i) | \mathbf{x}_i \in \mathcal{I}_t, \hat{y}_i \in \mathcal{C}_t \}, \nonumber\\
    \mathcal{S}_t^A = \{(\mathbf{x}_i', y_i') | (\mathbf{x}_i', y_i') \in \mathcal{S}, y_i' \in \mathcal{C}_t^A\}. \label{eq:active}
\end{gather}

\subsection{Efficient On-Device Dataset Condensation}\label{sec:condensation}

After assigning pseudo-labels to the $t$-th segment of the data stream and performing majority voting, we design efficient gradient matching to condense the data into the buffer. We employ a confidence-weighted cross-entropy loss as the model’s learning objective. We denote $\mathcal{X}$ and $\mathcal{Y}$ as the general image set and label set, respectively. Thus, $\mathcal{X}_t$ and $\mathcal{\hat{Y}}_t$ represent these sets for active incoming data $\mathcal{I}_t^A$, while $\mathcal{X}_t'$ and $\mathcal{Y}_t'$ represent the sets for the active synthetic data $\mathcal{S}_t^A$. The loss function is defined as follows:
\begin{equation}
    \mathcal{L}_{\theta}(\mathcal{X}, \mathcal{Y}) = - \sum_{i=1}^{|\mathcal{X}|} w_i \sum_{c \in \mathcal{C}} y_{i,c}  \log p_{\theta}(\mathbf{x}_i)_c,
\end{equation}
where $y_{i,c}$ equals 1 if $y_i = c$, and 0 otherwise. The weight $w_i$ is set to 1 for synthetic data $\mathcal{X}_t', \mathcal{Y}_t'$. For real data $\mathcal{X}_t$ and $\mathcal{\hat{Y}}_t$, however, $w_i$ is set as the confidence scores associated with them when generating the pseudo labels, i.e., $p_{\theta}(\mathbf{x}_i)_{\hat{y}_i}$. We designed this to prioritize higher confidence labels, which are likely to align more closely with the correct classifications.

The vanilla gradient matching framework, i.e., Eq.~\eqref{eq:opt1}, is a two-level optimization problem, requiring updates to synthetic images $\mathcal{S}$ in the outer loop and optimization of network parameters $\theta_t$ in the inner loop. This nested optimization process is computationally expensive and time-consuming for on-device learning. Thus, we consider simplifying the gradient matching process for better efficiency. Theoretically inspired by \cite{jin2022condensing}, we noted that the outer loop plays a more crucial role than the inner loop in gradient matching. From our empirical research, we observed two main outcomes: (1) a significant reduction in gradient matching loss immediately after model initialization, and (2) using multiple randomized models for a single step of gradient matching produced markedly better results than using one model for multiple steps of gradient matching, as shown in Fig.~\ref{fig:training_loss} of our experiment. Consequently, we introduce a simplified, one-step gradient matching strategy to speed up the condensation process. In this scheme, we focus solely on matching the gradients during the first epoch immediately after model initialization, while disregarding the training trajectory. Therefore, we can adjust our objective function by omitting $\sum_{t=0}^{T-1}$ as seen in Eq.~\eqref{eq:opt1}, thereby easing the constraints imposed on model training trajectories. We denote the initial randomized model parameters as $\tilde{\theta}$. The new objective for efficient gradient matching is:
\begin{equation}
    \argmin_{\mathcal{X}_t'} \mathcal{D}(\nabla_{\tilde{\theta}} \mathcal{L}_{\tilde{\theta}}(\mathcal{X}_t', \mathcal{Y}_t'), \nabla_{\tilde{\theta}} \mathcal{L}_{\tilde{\theta}}(\mathcal{X}_t, \mathcal{\hat{Y}}_t)).
    \label{eq:opt2}
\end{equation}

Furthermore, computing the value of Eq.~\eqref{eq:opt2} requires a second-order derivative of $\mathcal{D}$ with respect to $\mathcal{X}_t'$, which is computationally expensive. 
For simplicity, we denote $g_\text{syn} = \nabla_{\tilde{\theta}} \mathcal{L}_{\tilde{\theta}}(\mathcal{X}_t', \mathcal{Y}_t')$, $g_\text{real} = \nabla_{\tilde{\theta}} \mathcal{L}_{\tilde{\theta}}(\mathcal{X}_t, \mathcal{\hat{Y}}_t)$. Applying the chain rule to compute the gradient of the synthetic data yields
\begin{equation}
\begin{split}
     \nabla_{\mathcal{X}_t'} \mathcal{D}(g_\text{syn}, g_\text{real}) = \nabla_{g_\text{syn}} \mathcal{D}(g_\text{syn}, g_\text{real}) \cdot \nabla_{\mathcal{X}_t'} g_\text{syn} \\ = \nabla_{g_\text{syn}} \mathcal{D}(g_\text{syn}, g_\text{real}) \cdot \nabla_{\mathcal{X}_t'} \nabla_{\tilde{\theta}} \mathcal{L}_{\tilde{\theta}}(\mathcal{X}_t', \mathcal{Y}_t').
     \label{eq:opt3}
\end{split}
\end{equation}
This process is computationally demanding, as it contains an expensive matrix-vector product. Fortunately, the complexity can be substantially reduced using finite difference approximation. With $\epsilon$ as a small scalar\footnote{We used $\epsilon = 0.01 / ||\nabla_{g_\text{syn}} \mathcal{D}(g_\text{syn}, g_\text{real})||_2$ in our experiments, as suggested in previous work~\cite{liu2018darts}, and found it to be sufficiently accurate.} and $\tilde{\theta}^\pm = \tilde{\theta} \pm \epsilon \cdot \nabla_{g_\text{syn}} \mathcal{D}(g_\text{syn}, g_\text{real})$ we have:
\begin{equation}
    \nabla_{\mathcal{X}_t'} \mathcal{D}(g_\text{syn}, g_\text{real}) \approx \frac{1}{2 \epsilon} \left( \nabla_{\mathcal{X}_t'} \mathcal{L}_{\tilde{\theta}^+}(\mathcal{X}_t', \mathcal{Y}_t') - \nabla_{\mathcal{X}_t'} \mathcal{L}_{\tilde{\theta}^-}(\mathcal{X}_t', \mathcal{Y}_t') \right).
    \label{eq:opt4}
\end{equation}

With this approximation, we require five forward-backward passes in total to compute the gradient of the synthetic data $\mathcal{X}_t'$ with respect to the distance metric $\mathcal{D}$: the computation of $g_\text{syn}, g_\text{real}$, $\nabla_{g_\text{syn}} \mathcal{D}(g_\text{syn}, g_\text{real})$, and two terms in Eq.~\eqref{eq:opt4}. 
The computational efficiency is further enhanced as both the time and space complexity of our method are reduced from $\mathcal{O}(|\tilde{\theta}| \cdot |\mathcal{X}_t'|)$ to $\mathcal{O}(|\tilde{\theta}| + |\mathcal{X}_t'|)$.

\subsection{Enhancing Class Purity through Contrastive Learning}\label{sec:contrastive_learning}

\begin{figure}[h]
    \centering
    \includegraphics[width=0.48\textwidth]{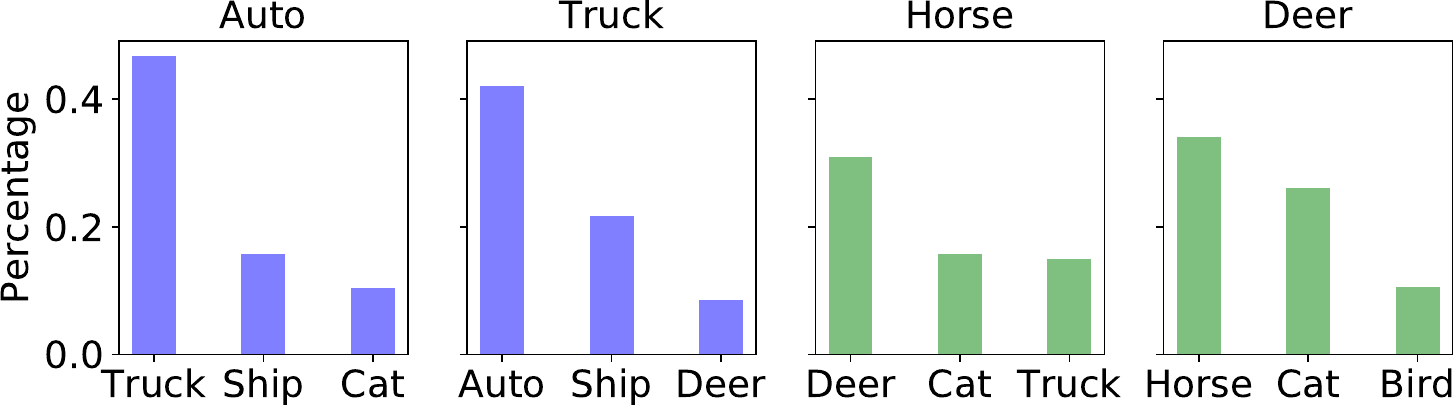}
    \caption{Top 3 classes most frequently misclassified in CIFAR-10 for selected classes.}
    \label{fig:confusion matrix}
\end{figure}

The condensation process can efficiently distill images into their corresponding classes. However, since the class of our incoming data is determined by the pre-trained model, the pseudo-labels may not always be accurate. Fig.~\ref{fig:confusion matrix} displays the top three classes most frequently misclassified within the CIFAR-10 dataset for several classes, with the y-axis representing their respective proportions of all misclassifications. Notably, classes with similar features, such as ``auto'' and ``truck'' or ``horse'' and ``deer'', are frequently confused with each other. As a result, incorrect pseudo-labels may cause images to be mistakenly grouped with other classes that share similar features. Over time, this misclassification can cause the synthetic data for similar classes increasingly alike, ultimately reducing the model's training effectiveness. Therefore, we need a method to minimize the impact of such ``noisy data'' within the buffer.

\begin{algorithm}
\caption{The proposed \name{} algorithm}
\label{algorithm}
\begin{algorithmic} 
\STATE \hskip\dimexpr-\algorithmicindent \textbf{Input:} input data stream $\mathcal{I}$, initial model parameters $\theta_0$, iteration number $L$, model optimizer $\text{opt}_\theta$, synthetic data optimizer $\text{opt}_\mathcal{S}$, hyper-parameters $\tau, \alpha$, $\beta$
\STATE \hskip\dimexpr-\algorithmicindent \textbf{Output:} condensed dataset $\mathcal{S}$, updated model parameters $\theta$
\STATE Deploy pre-trained model $\theta \leftarrow \theta_0$
\STATE Initialize condensed dataset $\mathcal{S}$ in buffer
\FOR{$\mathcal{I}_t \in \mathcal{I}$}
\STATE Assign pseudo-labels $\hat{y}_i$ for each $\mathbf{x}_i \in \mathcal{I}_t$
\STATE Identify active classes $\mathcal{C}_t^A$ from majority voting \hfill $\triangleright$ Eq.~\eqref{eq:filter}
\STATE Filter active streaming and synthetic data $\mathcal{I}_t^A, \mathcal{S}_t^A$ \hfill $\triangleright$ Eq.~\eqref{eq:active}
\FOR{$l \leftarrow 1$ to $L$}
    \STATE Randomize initial model parameters $\tilde{\theta}$
    \STATE Compute model gradients $g_{\text{syn}}, g_{\text{real}}$
    \STATE Perform efficient gradient matching for $\nabla_{\mathcal{X}_t'} \mathcal{D}$ \hfill $\triangleright$ Eq.~\eqref{eq:opt4}
    \STATE Compute contrastive learning loss $\mathcal{L}_{\text{cont}, \mathcal{S}}$ \hfill $\triangleright$ Eq.~\eqref{eq:contrast}
    \STATE Update condensed dataset $\mathcal{S}$ with $\text{opt}_\mathcal{S}$ \hfill $\triangleright$ Eq.~\eqref{eq:final}
\ENDFOR
\IF {$t\ \%\ \beta = 0$}
\STATE Update deployed model $\theta \leftarrow \text{opt}_{\theta}(\theta, \mathcal{S})$
\ENDIF
\ENDFOR
\end{algorithmic}
\end{algorithm}
\looseness=-1
Contrastive learning has been shown to enhance both the accuracy and robustness of classifiers. It refines the embedding space by increasing similarities within the same class and dissimilarities between different classes. This approach is particularly robust against the misattribution of pseudo-labels~\cite{khosla2020supervised}. We aim to utilize the contrastive learning objective to reduce the impact of mislabeled data.

Since each synthetic image in the buffer has a label, we utilize it as a constraint to design the contrastive learning loss. Specifically, for each active sample in $\mathcal{S}_t^A$, we found its corresponding index $i$ in $\mathcal{S}$, i.e., $(\mathbf{x}_i', y_i') \in \mathcal{S}$. Similarly, the set of indices of all current active samples in $\mathcal{S}$ is denoted as $A$. We consider all samples of class $y_i'$ except for itself as the positive samples, with their indices denoted by $P(i) = \{j | (\mathbf{x}_j, y_j) \in \mathcal{S}, y_j' = y_i', j \neq i\}$. We then randomly select a class different from $y_i$ as the negative class, denoted by $c^{\text{neg}}_i \in \mathcal{C}, c^{\text{neg}}_i \neq y_i'$. We consider all samples of class $c_i^{\text{neg}}$ as negative samples, with their indices represented by $N(i) = \{j | (\mathbf{x}_j, y_j) \in \mathcal{S}, y_j' = c_i^{\text{neg}}\}$. Assuming $f_\theta$ is the encoder of the deployed model $\theta$, we denote $z_i' = f_\theta(\mathbf{x}_i')$ as the feature representation of $\mathbf{x}_i' \in \mathcal{S}$. The contrastive learning loss is defined as follows:
\begin{gather}
    \mathcal{L}_{\text{cont},\mathcal{S}} = \sum_{i \in A} \frac{-1}{|P(i)|} \sum_{p \in P(i)} \log \frac{\exp(z_i' \cdot z_p' / \tau)}{\sum_{n \in N(i)} \exp(z_i' \cdot z_n' / \tau)},
    \label{eq:contrast}
\end{gather}
where $\cdot$ denotes the inner (dot) product, and $\tau$ denotes the temperature. By doing this, the model learns closely aligned representations for all samples from the same class, while pushing apart representations from different classes. 

\begin{table*}[tbp]
\small
\caption{Comparison of final average accuracy. ``Improvement'' indicates the increase over the best baseline.}

\centering
\begin{tabular}{cccccccccc}
\toprule
 & Labeled Ratio & IpC &  Random & FIFO & Selective-BP & K-Center & GSS-Greedy  & DECO (Ours) & Improvement\\ \midrule
\multirow{8}{*}{SVHN} & \multirow{4}{*}{10\%} & 1  & 51.20$\pm$1.23 & 55.70$\pm$1.01 & 52.61$\pm$1.15 & 51.50$\pm$1.02 & 53.37$\pm$0.99 & \textbf{71.18$\pm$0.09} & 27.8\%$\uparrow$ \\
 && 5 & 62.68$\pm$1.82 & 63.04$\pm$1.01 & 63.49$\pm$0.99 & 62.10$\pm$1.16 & 64.02$\pm$1.26 & \textbf{72.64$\pm$0.20} & 13.5\%$\uparrow$ \\
 && 10 & 74.52$\pm$1.59 & 73.45$\pm$1.66 & 70.60$\pm$1.23 & 70.25$\pm$1.32 & 72.16$\pm$1.45 & \textbf{75.89$\pm$0.16} & 1.2\%$\uparrow$ \\
 && 50 & 80.62$\pm$1.01 & 80.36$\pm$0.88 & 81.05$\pm$1.37 & 80.70$\pm$0.98 & 77.67$\pm$1.22 & \textbf{81.43$\pm$0.13} & 0.5\%$\uparrow$  \\
 \cmidrule(lr){2-10}
 &\multirow{4}{*}{1\%}& 1 & 46.31$\pm$1.23 & 39.39$\pm$1.18 & 47.43$\pm$1.32 & 45.89$\pm$1.67 & 44.32$\pm$1.52 & \textbf{65.84$\pm$0.19} &38.9\%$\uparrow$\\
 && 5 & 52.87$\pm$0.91 & 53.68$\pm$1.94 & 53.09$\pm$0.62 & 52.85$\pm$1.69 & 54.18$\pm$1.53 & \textbf{67.91$\pm$0.19} & 25.3\%$\uparrow$   \\
 && 10 & 64.21$\pm$1.77 & 61.95$\pm$1.15 & 67.79$\pm$1.60 & 65.48$\pm$1.22 & 63.18$\pm$0.89 & \textbf{74.29$\pm$0.35} & 9.6\%$\uparrow$  \\
 && 50 & 76.32$\pm$1.83 & 76.13$\pm$1.32 & 76.66$\pm$1.01 & 76.19$\pm$1.80 & 73.02$\pm$1.52 & \textbf{77.36$\pm$0.27 } & 0.9\%$\uparrow$ 
 \\ \midrule
\multirow{8}{*}{CIFAR-10} & \multirow{3}{*}{10\%} & 1  & 41.36$\pm$0.90 & 40.67$\pm$0.88 & 40.60$\pm$0.79 & 40.37$\pm$0.57 & 40.82$\pm$0.66 & \textbf{52.47$\pm$0.08} & 26.9\%$\uparrow$  \\
 && 5 & 42.18$\pm$1.01 & 44.82$\pm$0.75 & 42.93$\pm$1.40 & 43.26$\pm$0.66 & 45.60$\pm$0.88 & \textbf{57.02$\pm$0.13} & 25.0\%$\uparrow$  \\
 && 10  & 53.48$\pm$1.58 & 53.17$\pm$0.77 & 53.54$\pm$0.87 & 52.15$\pm$0.80 & 52.36$\pm$1.13 & \textbf{59.28$\pm$0.12} & 10.7\%$\uparrow$  \\
 && 50 & 58.46$\pm$0.66 & 59.80$\pm$0.28 & 60.60$\pm$0.40 & 58.98$\pm$0.59 & 59.87$\pm$1.01 & \textbf{62.41$\pm$0.22} & 3.0\%$\uparrow$   \\ 
  \cmidrule(lr){2-10}
 &\multirow{4}{*}{1\%}& 1 & 22.86$\pm$1.21 & 21.40$\pm$0.93 & 21.11$\pm$0.78 & 23.23$\pm$0.93 & 25.49$\pm$1.01 &\textbf{40.38$\pm$0.10} & 58.4\%$\uparrow$  \\
 && 5 & 29.50$\pm$0.88 & 31.28$\pm$0.61 & 30.11$\pm$1.12 & 30.80$\pm$0.69 & 31.66$\pm$0.90 & \textbf{47.78$\pm$0.07} & 50.9\%$\uparrow$  \\
 && 10 & 36.58$\pm$1.79 &40.15$\pm$1.22&38.48$\pm$0.66 &38.38$\pm$1.12 &39.84$\pm$1.08 & \textbf{48.90$\pm$0.08} & 21.8\%$\uparrow$ \\
 && 50 & 48.60$\pm$0.56 & 53.80$\pm$0.49 & 52.53$\pm$0.61 & 50.29$\pm$0.71 & 51.60$\pm$0.83 & \textbf{54.90$\pm$0.22} & 2.0\%$\uparrow$  
\\ \midrule
\multirow{8}{*}{CIFAR-100}& \multirow{3}{*}{20\%} & 1  & 18.93$\pm$0.79 & 18.42$\pm$0.57 & 16.82$\pm$0.48 & 18.26$\pm$0.44 & 17.46$\pm$0.33 & \textbf{22.24$\pm$0.06} & 17.5\%$\uparrow$  \\
 && 5 & 23.09$\pm$0.82 & 22.91$\pm$0.27 & 21.45$\pm$0.25 & 22.72$\pm$0.62 & 23.55$\pm$0.58 & \textbf{29.23$\pm$0.11} & 24.1\%$\uparrow$  \\
 && 10  & 26.23$\pm$0.48 & 26.40$\pm$0.52 & 25.80$\pm$0.23 & 25.91$\pm$0.20 & 25.97$\pm$0.40 & \textbf{33.01$\pm$0.19} & 25.0\%$\uparrow$   \\
 && 50 & 36.37$\pm$0.28 & 36.10$\pm$0.40 & 36.75$\pm$0.30 & 36.05$\pm$0.25 & 35.12$\pm$0.19 & \textbf{36.79$\pm$0.15} & 0.1\%$\uparrow$   \\ 
  \cmidrule(lr){2-10}
 &\multirow{4}{*}{10\%}& 1  & 10.15$\pm$0.32 & 10.07$\pm$0.22 & 12.01$\pm$0.28 & 12.13$\pm$0.23 & 14.16$\pm$0.48 & \textbf{22.06$\pm$0.05} & 55.8\%$\uparrow$   \\
 && 5 & 19.00$\pm$0.35 & 19.50$\pm$0.26 & 18.45$\pm$0.32 & 18.33$\pm$0.40 & 19.72$\pm$0.16 & \textbf{27.23$\pm$0.08} & 38.1\%$\uparrow$  \\
 && 10  & 16.91$\pm$0.51 & 21.59$\pm$0.42 & 21.71$\pm$0.18 & 20.60$\pm$0.21 & 21.65$\pm$0.36 & \textbf{29.01$\pm$0.15} & 33.6\%$\uparrow$  \\
 && 50 & 22.16$\pm$0.49 & 31.32$\pm$0.33 & 29.76$\pm$0.35 & 29.55$\pm$0.25 & 29.72$\pm$0.20 & \textbf{32.13$\pm$0.12} & 2.6\%$\uparrow$   \\ 
\midrule
\multirow{8}{*}{ImageNet-10}& \multirow{4}{*}{10\%} & 1 & 21.19$\pm$1.72 & 17.52$\pm$1.89 & 20.49$\pm$1.28 & 21.01$\pm$1.16 & 21.41$\pm$1.50 & \textbf{31.99$\pm$0.14} & 49.4\%$\uparrow$  \\
 && 5 & 32.98$\pm$2.47 & 32.46$\pm$2.30 & 32.83$\pm$0.94 & 32.80$\pm$1.33 & 32.36$\pm$1.59 & \textbf{41.02$\pm$0.23} & 24.4\%$\uparrow$  \\
 && 10 & 36.02$\pm$1.79 & 37.98$\pm$1.23 & 38.20$\pm$1.22 & 37.55$\pm$1.33 & 36.98$\pm$1.18 & \textbf{45.43$\pm$0.50} & 18.9\%$\uparrow$   \\
 && 50  & 45.20$\pm$1.10 & 54.43$\pm$1.33 & 52.66$\pm$1.35 & 50.02$\pm$0.72 & 50.65$\pm$0.86 & \textbf{59.42$\pm$0.23} & 9.2\%$\uparrow$   \\ 
  \cmidrule(lr){2-10}
 &\multirow{4}{*}{1\%}& 1 & 16.23$\pm$2.16 & 15.10$\pm$1.78 & 17.03$\pm$1.90 & 17.47$\pm$1.35 & 18.62$\pm$1.22 & \textbf{25.80$\pm$0.25}& 38.6\%$\uparrow$  \\
 && 5 & 17.96$\pm$3.77 & 19.22$\pm$2.38 & 20.24$\pm$2.91 & 20.48$\pm$1.25 & 20.34$\pm$1.90 & \textbf{28.29$\pm$0.19}  & 38.1\%$\uparrow$ \\
 && 10 & 20.67$\pm$2.10 & 22.48$\pm$0.98 & 22.99$\pm$1.39 & 22.38$\pm$1.30 & 21.28$\pm$1.45 & \textbf{29.58$\pm$0.09} & 28.7\%$\uparrow$  \\
 && 50 & 22.55$\pm$1.89 &23.01$\pm$2.21 & 23.51$\pm$0.99 & 23.35$\pm$0.59 & 22.83$\pm$1.20 & \textbf{31.55$\pm$0.20} & 34.2\%$\uparrow$ 
\\ 
\bottomrule
\end{tabular}
\label{tab: main}
\end{table*}

\subsection{Overall Optimization}

Eq.~\eqref{eq:final} below represents the overall optimization for $\mathcal{S}$. Note that the parameters of $\mathcal{X}_t'$ used in Section~\ref{sec:condensation} for $\mathcal{D}$ are a subset of $\mathcal{S}$.
\begin{equation}
    \text{opt}_{\mathcal{S}} \left( \nabla_{\mathcal{S}} \mathcal{D}(g_\text{syn}, g_\text{real}) + \alpha \cdot \nabla_{\mathcal{S}} \mathcal{L}_{\text{cont}, \mathcal{S}} \right).
    \label{eq:final}
\end{equation}
Here, \( \alpha \) serves as a weighting factor that balances the gradient matching loss with the contrastive learning loss. $\text{opt}_{\mathcal{S}}$ is an optimizer (e.g., SGD) to update the condensed dataset $\mathcal{S}$. After every $\beta$ segments streaming data, we use $\mathcal{S}$ to update the deployed model $\theta$ with $\text{opt}_\theta$. The whole process is shown in Algorithm~\ref{algorithm}.
\section{Experiments}

\subsection{Experimental Setups}
\subsubsection{Datasets and Evaluation Protocols}
We conduct experiments on SVHN \cite{netzer2011reading}, CIFAR-10, CIFAR-100 \cite{krizhevsky2009learning} and ImageNet-10 \cite{russakovsky2015imagenet} to evaluate our method. For SVHN, CIFAR-10, and CIFAR-100, we use the standard train-test splits. For ImageNet-10, we select the first 10 classes based on \cite{tian2020contrastive} and divide the dataset into training (80\%), validation (10\%), and testing (10\%) portions. To initialize the simulation, we pre-trained the models on datasets with varying labeled data ratios. Specifically, for SVHN, CIFAR-10, and ImageNet-10, the models are pre-trained using 10\% and 1\% labeled data. Given the larger number of classes in CIFAR-100 and the insufficient accuracy derived from 1\% labeled data, we opt to pre-train models on 20\% and 10\% labeled data. To mimic the temporal correlation present in data streams, we employ the Strength of Temporal Correlation (STC) \cite{wu2021enabling, hayes2019memory}. STC represents the number of consecutive data in the input stream belonging to the same class until a class change occurs, with a higher STC indicating stronger class consistency over time. STC is set to 500 for SVHN, CIFAR-10, CIFAR-100, and 100 for ImageNet-10.\looseness=-1

\subsubsection{Baselines}
Our approach is compared against five baseline methods (Random \cite{vitter1985random}, FIFO \cite{hayes2019memory}, Selective-BP \cite{jiang2019accelerating, killamsetty2021grad}, K-Center~\cite{lu2020semantic, sener2017active} and GSS-Greedy \cite{aljundi2019gradient, ha2023online}). \textit{Random} selects a random subset for the new data buffer. \textit{FIFO}, a method used in continual learning, replaces the oldest buffer data with new entries. \textit{Selective-BP} selects data with lower prediction confidence as determined by the model for storage in the buffer. \textit{K-Center} selects the centers that minimize the largest distance between a sample and its nearest center. \textit{GSS-Greedy} evaluates the similarity of buffer data, prioritizing the replacement of similar items with distinct new data.

\subsubsection{Implementation Details}
We use ConvNet \cite{gidaris2018dynamic} as the backbone model for all experiments, and employ SGD with momentum as the optimizer. Unless otherwise specified, the batch size is 128 with the weight decay 5e-4. We set the iteration number $L$ to 10, the filtering threshold $M$ to 0.4, the scalar temperature factor $\tau$ in the contrastive loss to 0.07, and the weight factor $alpha$ in the loss function to 0.1. We use the cosine similarity as the distance metric $\mathcal{D}$ for gradient matching. For SVHN, CIFAR-10, and CIFAR-100, the learning rate is 1e-3; for ImageNet-10, it is 1e-4. For all datasets, the training interval $\beta$ is set to 10, and the model is trained for 200 epochs on the condensed dataset for each update. The Images per Class (IpC) value of the condensed dataset in the buffer defaults to 10 unless otherwise specified, which also indicates the required buffer size.
For each experimental setup, we conduct five trials with different random seeds and report the average results and variance.

\subsection{Experimental Results}
\subsubsection{Classification Performance}
For classification performance comparison, we report the average end accuracy of our methods with variance compared to the baselines under different labeled ratios, which is presented in Table~\ref{tab: main}. 
To evaluate the effectiveness of our method across various sizes of condensed datasets (i.e., required buffer sizes), we present the results for different IpC values in the buffer, specifically \{1, 5, 10, 50\}. We have the following observations:

\begin{compactenum}[(a)]
    \item Our method consistently outperforms baseline methods across different labeled ratios, IpCs, and datasets. Notably, with a small IpC value, our method significantly surpasses the baselines. For instance, with IpC=1 and a labeled ratio of 0.1, our method exceeds the top baseline by 15.48\% on SVHN, 11.11\% on CIFAR-10, 3.31\% on CIFAR-100, and 10.58\% on ImageNet-10. This demonstrates our method's ability to leverage new data in scenarios with limited memory resources.
    \item Our method consistently demonstrates significantly lower variance compared to baseline methods across all experimental settings. This improvement is primarily because baseline methods must remove old data when new data arrives, which greatly increases uncertainty due to substantial buffer updates. In contrast, our method directly condenses new data into the buffer without removing old data, resulting in less randomness and enhanced stability in its execution.
    \item Our approach shows greater benefits with smaller labeled ratios. For example, on CIFAR-10 with IpC=1, reducing the labeled ratio from 0.1 to 0.01 results in a performance decrease of over 15\% for all baseline methods, while our method sees only a 12.09\% reduction. With fewer labeled data, our method more effectively utilizes new information for model updates.
    \item CIFAR-100's performance was not as strong as that on other datasets, likely due to its greater class count and data diversity. With the IpC value set to 50, a comparatively large data buffer with a size of 5000 is constructed. This expanded buffer capacity naturally reduces the relative advantage of our method, as the challenges associated with buffer capacity are mitigated. Despite this, our approach continues to outperform baselines.
\end{compactenum}

\begin{figure*}[ht]
    \centering
    \begin{subfigure}[b]{0.23\linewidth}
        \includegraphics[width=1\linewidth]{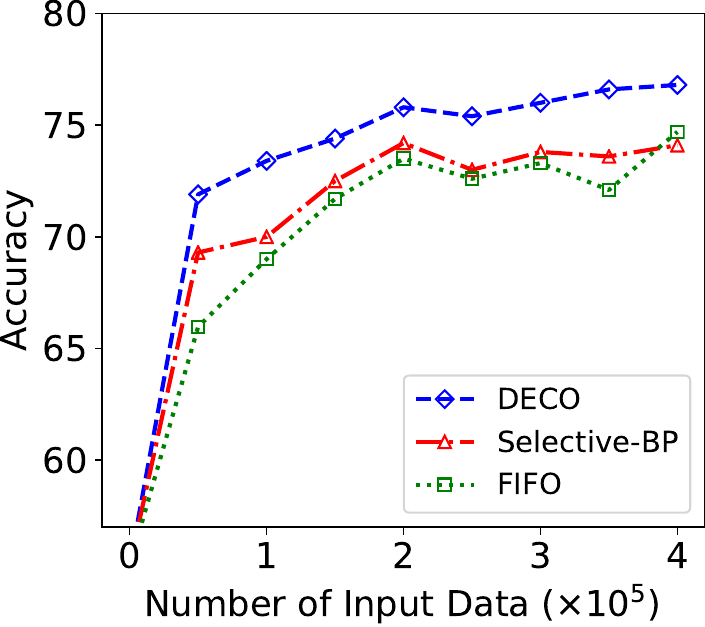}
        \caption{SVHN}
        \label{fig:training_curve_svhn}
    \end{subfigure}
    \hfill
    \begin{subfigure}[b]{0.23\linewidth}
        \includegraphics[width=1\linewidth]{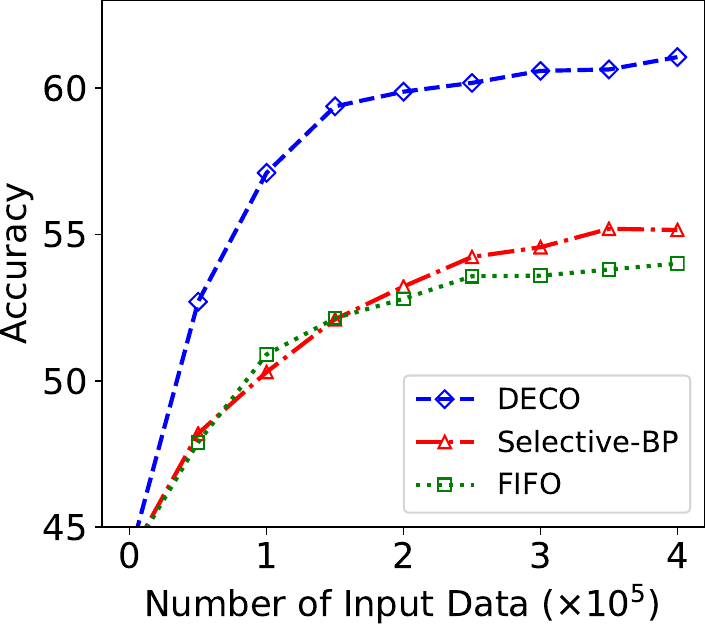}
        \caption{CIFAR-10}
        \label{fig:training_curve_CIFAR-10}
    \end{subfigure}
    \hfill
    \begin{subfigure}[b]{0.23\linewidth}
        \includegraphics[width=1\linewidth]{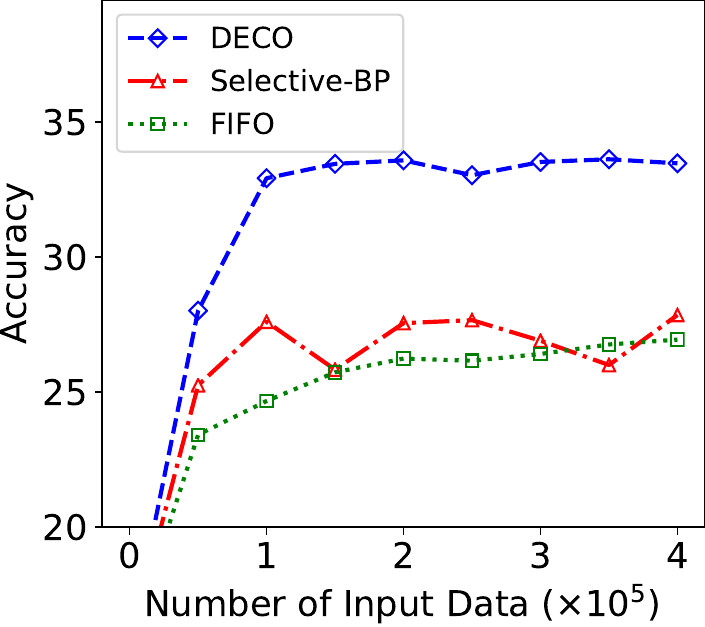}
        \caption{CIFAR-100}
        \label{fig:training_curve_CIFAR-100}
    \end{subfigure}
    \hfill
        \begin{subfigure}[b]{0.23\linewidth}
        \includegraphics[width=1\linewidth]{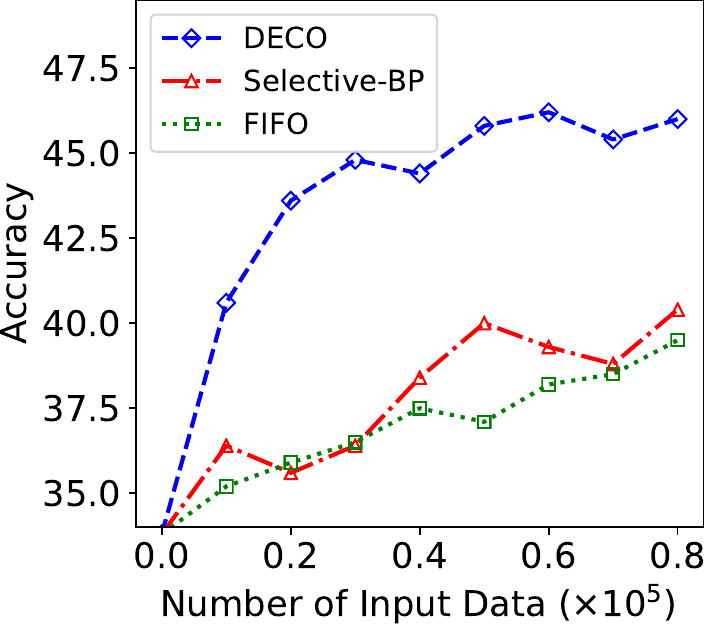}
        \caption{ImageNet-10}
        \label{fig:training_curve_ImageNet-10}
    \end{subfigure}
    \vskip -0.5em
    \caption{Learning curves on different datasets depict the average accuracy in relation to the amount of input data.}
    \label{fig:learning_curve}

\end{figure*}

\begin{figure*}[ht]
    \centering
    \begin{subfigure}[b]{0.23\linewidth}
        \includegraphics[width=1\linewidth]{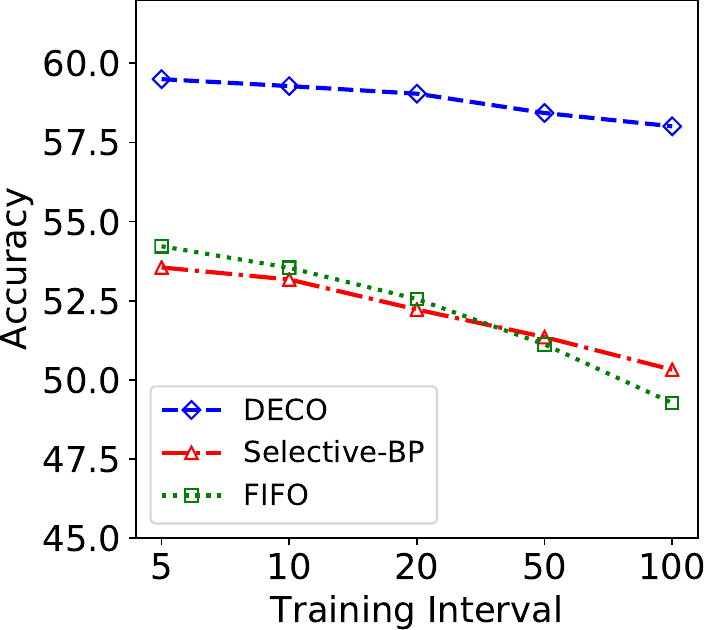}
        \caption{Training intervals $\beta$}
        \label{fig:training_interval}
    \end{subfigure}
    \hfill
    \begin{subfigure}[b]{0.23\linewidth}
        \includegraphics[width=1\linewidth]{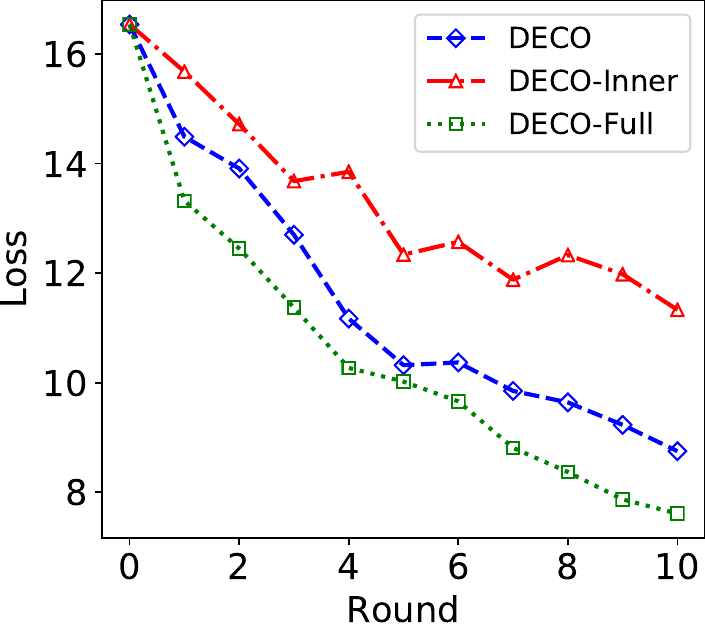}
        \caption{Condensation settings}
        \label{fig:training_loss}
    \end{subfigure}
    \hfill
    \begin{subfigure}[b]{0.23\linewidth}
        \includegraphics[width=1\linewidth]{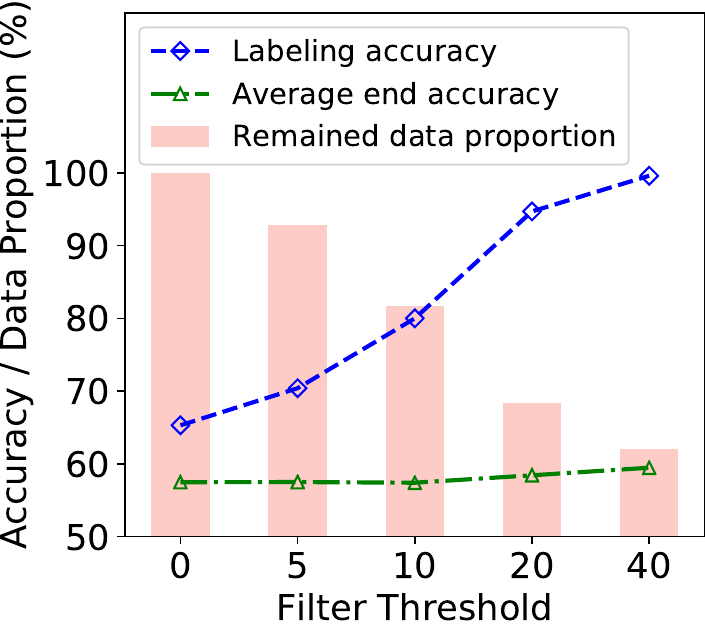}
        \caption{Filter threshold $M$}
        \label{fig:filter_size}
    \end{subfigure}
    \hfill
    \begin{subfigure}[b]{0.23\linewidth}
        \includegraphics[width=1\linewidth]{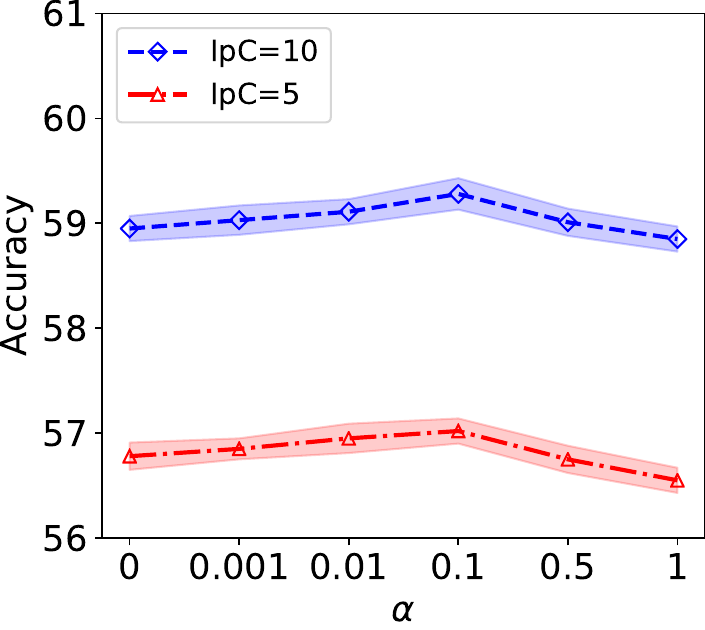}
        \caption{Weighting factor $\alpha$}
        \label{fig:alpha}
    \end{subfigure}
    \vskip -0.5em
    \caption{Parameter analysis. (a) shows the effect of varying training intervals $\beta$ on model performance. (b) shows the loss curve for different condensation optimization settings. (c) shows the influence of different filter thresholds $M$ on pseudo-labeling accuracy and classification performance. (d) shows the impact of the loss weighting factor $\alpha$ on average end accuracy.}
\end{figure*}

\subsubsection{Training Curve} \label{sec: training_surve}
In this section, we analyze our model's training efficiency by comparing its learning curve on the datasets with the two most competitive baselines, FIFO and Selective-BP. The learning curve shows the speed at which our model learns from new data. This comparison is depicted in Fig.~\ref{fig:learning_curve}, where the x-axis represents the number of processed inputs and the y-axis indicates model accuracy. Our observations show that our method consistently achieves higher accuracy than the baseline.

For the SVHN dataset, which has relatively easier classification tasks, the difference between our method and the baseline is not very pronounced. However, our method still surpasses the baseline's final model accuracy by 1.1\% after processing only 50\% of the input data, demonstrating its efficiency in learning new information. The advantage of our method is more significant in the other three datasets. Specifically, for the CIFAR-10 dataset, after processing only 100,000 input data points, our model's accuracy escalates to 57.1\%, a benchmark the baseline struggles to meet even after analyzing the entire dataset. For the CIFAR-100 and ImageNet-10 datasets, our method outperforms the baseline's final model accuracy using just an eighth of the data. By the end of the model training, it exceeds the best baseline models by 5.61\% and 5.60\% accuracy, respectively. This indicates that the baseline methods lose a substantial amount of information by removing old data, highlighting the advantages of condensation methods. Additionally, we observed that the learning curve of our method is smoother across all datasets compared to the baseline methods, which shows that our method is more robust in the learning process.

\begin{table}[htbp]
\small
\caption{Comparison of execution time.}
\begin{tabular}{cccccc}
\toprule
Dataset & Method & IpC=1 & IpC=10 & IpC=50 \\ 
\midrule
\multirow{3}{*}{CIFAR-10} & Selective-BP & 2.3 (40.6) & 3.5 (53.5)& 11.7 (60.6) \\
 & DECO-Full & 54.0 (51.6) & 139.8 (60.2)& 180.5 (62.4)\\
 & \name{} & 7.5 (52.5)& 15.9 (59.3)& 45.4 (62.4)\\
\midrule
\multirow{3}{*}{ImageNet-10} & Selective-BP & 5.2 (20.5)& 24.2 (38.2) & 66.1 (52.7)\\
 & DECO-Full & 168.1 (33.0)  & 894.2 (44.3)& 2232.9 (59.0)\\
 & \name{} & 22.2 (32.0)& 71.6 (45.4)& 202.8 (59.4)\\
\bottomrule
\end{tabular}
\label{tab:running_time}
\end{table}

\subsubsection{Time Complexity Analysis}
We analyzed the efficiency of our methods by measuring the total execution time on the CIFAR-10 dataset and ImageNet-10 dataset, the result is shown in Table~\ref{tab:running_time}. We compared our method against the most effective baseline method, Selective-BP, and the condensation approach without any acceleration and optimization, noticed as DECO-Full. To facilitate direct comparisons, we included the average end accuracy for each setting in parentheses next to the corresponding execution times. Our findings are shown below:

\begin{compactenum}[(a)]
    \item Compared to DECO-Full, our optimization significantly enhances efficiency. Specifically, on the CIFAR-10 dataset, our method achieves speed improvements of 3.3x, 8.7x, 8.8x, and 4x at IpC values of 1, 5, 10, and 50, respectively. This acceleration is even more pronounced on ImageNet-10 datasets with larger image sizes. Moreover, we have found that our optimization method not only improves efficiency but also does not significantly decrease accuracy. In some cases, the accuracy even increases. Moreover, the enhanced efficiency of our method does not result in significant accuracy loss; in fact, accuracy slightly improves in some cases. This underscores the effectiveness of our optimizations to the condensation algorithm.
    \item Although the total condensation time increases with the buffer size, the rate of increase is proportionally smaller than the growth of the buffer size itself. For example, the total training time extends by a factor of 6.1x as the IpC increases from 1 to 50 in the CIFAR-10 dataset. This increment is comparable to the Selective-BP methods, which experience a 5.1x increase. Consequently, the processing time of our method remains manageable despite the increase in buffer size, demonstrating its scalability to larger buffer configurations.
    \item The time our method requires consistently remains within five times that of the Selective-BP method. Although our approach takes longer compared to selection-based methods, the training curve in Section~\ref{sec: training_surve} demonstrates that our method can achieve comparable or even better performance with substantially less data than Selective-BP requires. Therefore, we consider this compromise acceptable for on-device learning.
\end{compactenum}
\subsubsection{Evaluation of Training Intervals $\beta$}
We study how varying training intervals influence model performance. Training intervals determine the frequency of model updates from the buffer, impacting both computational load and accuracy. Our results, shown in Fig.~\ref{fig:training_interval}, indicate that as training intervals increase, the accuracy of the baseline method decreases significantly. In contrast, our method shows only a slight and gradual decline in accuracy, indicating a more stable performance trend overall. This stability results from our method's ability to condense new incoming data into the buffer without discarding old data samples. Consequently, even with longer training intervals, our buffer preserves more information, which helps maintain model accuracy.  
This characteristic is beneficial as it allows for more flexible model training by decreasing the frequency of updates, thus lowering computational costs.

\subsubsection{Evaluation of Condensation Settings}

To enhance the efficiency of the condensation process, we removed the inner loop as described in Section~\ref{sec:condensation}. To better understand the roles of the two loops in the condensation process, we plotted the loss curves under three different settings: only the outer loop (DECO), only the inner loop (DECO-Inner), and both loops (DECO-Full), as shown in Fig.~\ref{fig:training_loss}. The loss decreases more slowly under the DECO-Inner setting compared to the original DECO. Additionally, DECO and DECO-Full have a relatively small gap between them. Given that the experimental results in Table~\ref{tab:running_time} indicate that each round's duration in DECO-Full is significantly longer than in DECO, we believe that omitting the inner loop appears to be a more reasonable approach.

\subsubsection{Evaluation of Filter Threshold $M$}

Fig.~\ref{fig:filter_size} demonstrates the impact of the filter threshold $M$ on three key metrics: the percentage of data retained after filtering, the accuracy of the generated pseudo-labels, and the overall model accuracy. The x-axis represents the filter threshold, while the y-axis shows both accuracy and retained data proportion. As the filter threshold increases, less data meets this threshold, but the accuracy of the pseudo-labels for the remaining data improves. Specifically, with no filtering at a threshold of 0, the pseudo-labeling accuracy is only 65.3\%. However, increasing the threshold to 40 results in only 62\% of the data being retained, yet the accuracy of the pseudo-labels jumps to 99.6\%. This illustrates a trade-off between the amount of data retained and the quality of the labels. The data indicates that the highest training accuracy occurs at a filter threshold of 40, suggesting that label accuracy is more crucial than data volume for effective training.

\subsubsection{Evaluation of Weighting Factor $\alpha$}
A parameter analysis was conducted to assess the impact of the contrastive loss term on performance; the results are shown in Fig.~\ref{fig:alpha}. We varied the coefficient \(\alpha\) within the range of {0, 0.00, 0.01, 0.1, 0.5, 1}, and reported the average end accuracy of classification on CIFAR-10 for IpC at 5 and 10. The findings indicate that for both IpC settings, accuracy improves as \(\alpha\) increases from 0 to 0.1, with the optimal setting being \(\alpha=0.1\). This demonstrates that incorporating an appropriate amount of contrastive loss can enhance model performance.
\section{Conclusion}

This study aims to enhance the learning process of deployed models in edge-device environments through dataset condensation techniques. We introduce a framework to manage a limited-size data buffer containing synthetic data. Initially, pseudo-labels are assigned to streaming data with majority voting. Subsequently, an efficient gradient matching technique is employed to condense this data into its respective classes. Moreover, to mitigate the impact of labeling noise, we incorporate a contrastive learning objective to improve the quality of the buffered data. Our results demonstrate significantly improved performance compared to existing methods.

\bibliographystyle{ACM-Reference-Format}
\bibliography{sample}

\end{document}